\documentclass[lettersize,journal]{IEEEtran}
\usepackage{amsmath,amsfonts}
\usepackage{amssymb}
\usepackage{algorithm}
\usepackage{latexsym}
\usepackage{array}
\usepackage[caption=false,font=normalsize,labelfont=sf,textfont=sf]{subfig}
\usepackage{xcolor}
\usepackage{textcomp}
\usepackage[noend]{algpseudocode}
\usepackage{microtype}
\usepackage{tabularray}
\usepackage{hyperref}
\usepackage{stfloats}
\usepackage{url}
\usepackage{booktabs}
\usepackage{verbatim}
\usepackage{graphicx}
\usepackage{cite}
\begin{document}

\title{ONG: One-Shot NMF-based Gradient Masking for Efficient Model Sparsification}


\author{\IEEEauthorblockN{Sankar Behera, and Yamuna Prasad}

\IEEEauthorblockA{
Dept. of CSE, Indian Institute of Technology Jammu, India; \{sankar.behera, yamuna.prasad\}@iitjammu.ac.in
}

}



\maketitle
\begin{abstract}Deep Neural Networks (DNNs) have achieved remarkable success but their large size poses deployment challenges. While various pruning techniques exist, many involve complex iterative processes, specialized criteria, or struggle to maintain sparsity effectively during training. We introduce ONG (One-shot NMF-based Gradient Masking), a novel sparsification strategy that identifies salient weight structures using Non-negative Matrix Factorization (NMF) for one-shot pruning at the outset of training. Subsequently, ONG employs a precise gradient masking mechanism to ensure that only unpruned weights are updated, strictly preserving the target sparsity throughout the training phase. We integrate ONG into the BIMP comparative framework and evaluate it on CIFAR-10 and CIFAR-100 with ResNet56, ResNet34, and ResNet18 against established stable sparsification methods. Our experiments demonstrate ONG's ability to achieve comparable or superior performance at various sparsity levels while maintaining structural integrity post-pruning and offering a clear mechanism for targeting desired sparsities.
\end{abstract}


\section{Introduction}
The remarkable advancements in deep neural networks (DNNs) have revolutionized fields ranging from natural language processing to computer vision \cite{vaswani2017attention,krizhevsky2012imagenet,he2016deep}. This success, however, is often coupled with a dramatic increase in model size and computational demands, epitomized by Large Language Models (LLMs) with billions of parameters \cite{kaplan2020scaling,touvron2023llama}. Such scale poses significant challenges for deployment, particularly on resource-constrained environments like mobile devices or edge hardware, limiting their accessibility and practical application. Consequently, model compression has emerged as a critical area of research, aiming to reduce the memory footprint and inference latency of large models without substantially compromising their performance. Key compression paradigms include quantization \cite{jacob2018quantization}, knowledge distillation \cite{hinton2015distilling}, low-rank decomposition \cite{denton2014exploiting}, and network pruning.

Network pruning, the focus of this work, seeks to eliminate redundant parameters or structural components from over-parameterized networks. Pruning techniques are broadly classified into unstructured pruning, which removes individual weights leading to irregular sparsity patterns often requiring specialized hardware for acceleration \cite{han2015learning}, and structured pruning, which removes entire channels, filters, or even layers, generally offering better compatibility with existing hardware \cite{li2017pruning,he2017channel}. While structured pruning can be more hardware-friendly, it may discard entire blocks of computation that contain a mix of important and unimportant information, potentially leading to a more significant accuracy drop for a given parameter reduction compared to fine-grained unstructured pruning. Intra-channel pruning methods \cite{wen2016learning,park2023dynamic,mao2017exploring} attempt to strike a balance by operating at a granularity finer than entire channels but coarser than individual weights.

The \textit{timing} of pruning also differentiates various approaches. Traditional methods often involve a three-stage pipeline: pretraining a dense model, pruning it based on some criteria (e.g., weight magnitude), and then fine-tuning the sparse model to recover performance \cite{han2015learning}. While effective, this iterative process can be computationally intensive. More recent efforts have focused on integrating pruning directly into the training process ("pruning during training") or even determining sparse architectures before training begins ("pruning at initialization"). Dynamic pruning methods \cite{zhu2017to,lin2020dynamic,bellec2018deep} adjust the sparsity pattern throughout training, often guided by heuristics, gradient information, or error feedback mechanisms. Other "stable" strategies aim to learn sparse models from scratch by reparameterizing weights or introducing sparsity-inducing regularizers \cite{kusupati2020soft,savarese2020winning,liu2021do}. Pruning at initialization, spurred by findings like the Lottery Ticket Hypothesis \cite{frankle2019lottery}, seeks to identify highly trainable sparse subnetworks from the outset \cite{lee2019snip}.

A persistent challenge across many pruning-during-training schemes is the maintenance of the achieved sparsity. Standard optimizer updates, particularly those involving momentum or weight decay, can inadvertently cause previously pruned (zeroed-out) weights to become non-zero again, leading to a degradation of the intended sparsity structure over time if not explicitly counteracted. This necessitates careful handling of gradients and weight updates for the pruned elements.

In this paper, we introduce ONG (One-shot NMF-based Gradient Masking), a novel sparsification strategy designed to address these challenges. ONG uniquely combines the efficiency of one-shot pruning with the stability required for training a fixed sparse architecture. It leverages Non-negative Matrix Factorization (NMF) as a data-driven approach to assess the structural importance of weights based on their reconstruction error from a low-rank basis. This NMF-derived importance score guides a single, decisive pruning operation performed \textit{before} training commences. Subsequently, during the training phase, ONG employs a meticulous and strict gradient and weight masking mechanism. This ensures that only the weights corresponding to the initially determined unpruned mask are allowed to receive updates, thereby strictly preserving the target sparsity level throughout the entire training process without degradation. ONG further incorporates a mechanism to automatically tune its pruning threshold to achieve a user-specified global sparsity target.

Our contributions are thus:
\begin{enumerate}
    \item A novel weight importance scoring mechanism based on NMF reconstruction error, providing a structural perspective on weight saliency beyond simple magnitude.
    \item The ONG strategy, which integrates one-shot NMF-driven pruning with a robust training phase characterized by strict gradient and weight masking to ensure unwavering sparsity.
    \item An automated method for determining the NMF pruning threshold to precisely achieve a desired global sparsity level.
    \item A comprehensive empirical evaluation of ONG within the BIMP (Behavioral Insights for Model Pruning) comparative framework \cite{zimmer2023how}. We benchmark ONG against a suite of established stable sparsification strategies using the CIFAR-10 dataset and ResNet56 architecture, demonstrating its effectiveness and competitive performance.
\end{enumerate}
The remainder of this paper is organized as follows: Section 2 reviews related work. Section 3 introduces preliminary concepts and notations. Section 4 details the ONG methodology. Section 5 presents our experimental setup and results. Finally, Section 6 concludes the paper and discusses future directions.

\section{Related Work}
Network pruning has a rich history, with numerous approaches proposed to reduce model complexity. We categorize relevant works based on their core mechanisms and pruning schedule.

\paragraph{Magnitude-based Pruning.} One of the earliest and most influential approaches is pruning weights with the smallest absolute magnitudes \cite{han2015learning}. This is often performed iteratively with fine-tuning. Gradual Magnitude Pruning (GMP) \cite{zhu2017to} extends this by progressively increasing the sparsity ratio during a single training run, avoiding full retraining cycles. While simple and effective, magnitude alone may not always capture the functional importance of a weight within the network structure.

\paragraph{Sparsity-Inducing Regularization and Reparameterization.} Many "stable" strategies train sparse models from scratch by modifying the learning objective or weight representations.
STR (Soft Threshold Reparameterization) \cite{kusupati2020soft} introduces a learnable soft threshold. CS (Continuous Sparsification) \cite{savarese2020winning} uses learnable gates with regularization and annealing. DST (Dynamic Sparse Training) \cite{liu2021do} learns per-channel/filter thresholds. LC (``Learning Compression'') \cite{carreira2018learning} modifies weight decay. GSM (Global Sparse Momentum) \cite{ding2019global} applies sparsity to momentum updates. These methods typically converge to a sparse model but may require careful tuning.

\paragraph{Dynamic Pruning and Error Correction.} Some methods dynamically adjust the sparsity mask during training. DPF (Dynamic Model Pruning with Feedback) \cite{lin2020dynamic} maintains a dense model alongside the sparse one, using error feedback for recovery from premature pruning. Other methods like DeepR \cite{bellec2018deep} and SET \cite{mocanu2018scalable} explore dynamic rewiring, often using heuristics for regrowth.

\paragraph{Pruning at Initialization.} Inspired by the Lottery Ticket Hypothesis \cite{frankle2019lottery}, which suggests that dense networks contain sparse subnetworks (winning tickets) that can be trained in isolation to achieve comparable performance, several methods aim to identify such subnetworks at or near initialization. SNIP \cite{lee2019snip} uses connection sensitivity, a measure based on how much the loss changes with respect to removing a connection, to prune weights before training. Other methods explore gradient-based signals \cite{wang2020picking}.

\paragraph{Low-Rank Decomposition and Matrix Factorization.} These techniques approximate weight matrices with the product of smaller, lower-rank matrices \cite{denton2014exploiting,jaderberg2014speeding}. This is inherently a form of structured compression. Recent work like SoLA \cite{huang2025sola} combines low-rank decomposition with soft activation sparsity in LLMs. SVD-LLM \cite{wang2024svdllm} focuses on truncation-aware SVD. While ONG also employs a matrix factorization technique (NMF), its purpose is distinct: NMF is used to derive element-wise *importance scores* for selecting which original weights to keep or prune, rather than replacing matrices with their factorized counterparts.

\paragraph{NMF in Neural Networks.} NMF \cite{lee1999learning} has been used for various tasks such as dimensionality reduction, feature learning, and dictionary learning. In the context of neural networks, it has been explored for layer decomposition or learning parts-based representations. Its application as a primary criterion for deriving element-wise pruning masks based on reconstruction error, followed by stable sparse training with strict gradient and weight enforcement, as proposed in ONG, is a novel combination. ONG aims to leverage NMF's ability to capture latent structure to make more informed pruning decisions than magnitude alone, while maintaining the simplicity of one-shot mask determination.

Our proposed ONG method draws inspiration from one-shot pruning for its efficiency in mask determination and from stable sparse training methods for its robust training phase. However, it introduces NMF-based scoring as a novel criterion and emphasizes strict sparsity preservation through explicit gradient and weight masking, a feature crucial for predictable sparse model behavior.

\section{The ONG Method}
\label{sec:ong_method}


The ONG (One-shot NMF-based Gradient Masking) strategy is designed to efficiently create and train sparse neural networks. It operates in three primary stages, detailed below and summarized in Algorithm \ref{alg:ong}: (1) NMF-based weight importance scoring, (2) One-shot mask determination incorporating automated $\gamma$-tuning for target sparsity, and (3) Training the resultant sparse model with rigorous gradient and weight masking.

\subsection{NMF-based Weight Importance Scoring}
The foundational step in ONG is to assign an importance score to each weight, reflecting its structural contribution as identified by Non-negative Matrix Factorization (NMF). For every prunable weight matrix $\mathbf{W}$ in the network (typically from Conv2D or Linear layers, excluding layers like final classifiers or Batch Normalization), we perform the following:
\begin{enumerate}
    \item \textbf{Reshape and Absolutize:} The weight matrix $\mathbf{W} \in \mathbb{R}^{m \times n}$ (where $n$ can represent flattened subsequent dimensions for convolutional layers) is reshaped to $\mathbf{W}_{\text{2D}} \in \mathbb{R}^{m \times p}$. As NMF requires non-negative inputs, we use the absolute values: $\mathbf{W}_{\text{abs}} = |\mathbf{W}_{\text{2D}}|$.
    \item \textbf{NMF Decomposition:} $\mathbf{W}_{\text{abs}}$ is decomposed into two non-negative factor matrices, $\mathbf{F} \in \mathbb{R}^{m \times k}$ (basis matrix) and $\mathbf{G} \in \mathbb{R}^{k \times p}$ (coefficient matrix), where $k$ is the number of NMF components (a key hyperparameter of ONG). This decomposition aims to minimize the Frobenius norm of the reconstruction error:
    \begin{equation}
        \min_{\mathbf{F} \ge 0, \mathbf{G} \ge 0} \|\mathbf{W}_{\text{abs}} - \mathbf{F}\mathbf{G}\|_F^2
    \label{eq:nmf_objective}
    \end{equation}
    \item \textbf{Reconstruction and Score Calculation:} The NMF reconstruction is $\mathbf{\hat{W}}_{\text{abs}} = \mathbf{F} \mathbf{G}$. The importance score for each original element $w_{ij}$ (corresponding to $w_{\text{abs},ij}$) is its absolute reconstruction error (delta-based score):
    \begin{equation}
        Score(w_{ij}) = | w_{\text{abs},ij} - \hat{w}_{\text{abs},ij} |
    \label{eq:nmf_score}
    \end{equation}
    A higher score signifies that $w_{ij}$ is poorly approximated by the $k$-component NMF, suggesting it captures unique information not well represented by the dominant $k$ latent factors and is thus considered more important to retain. This scoring is performed once on the initial model weights, with $N_{iter}$ denoting the NMF optimization iterations.
\end{enumerate}

\subsection{One-Shot Mask Determination with Automated $\gamma$-Tuning}
\label{subsec:mask_determination_gamma_tuning}
With NMF-derived scores, ONG determines a binary mask $\mathbf{M}_L$ for each layer L. This process involves calculating a pruning threshold $\tau_L$, which is controlled by a statistical scaling factor $\gamma$. To achieve a user-defined global target sparsity $S_{\text{target}}$, ONG employs an automated binary search mechanism to find an optimal $\gamma^*$.

\begin{enumerate}
    \item \textbf{Statistical Threshold Calculation (per layer L):} Given a scaling factor $\gamma$ and the score matrix $Score_L$ for layer L, one of two statistical thresholding strategies (a hyperparameter $T_{type}$) is applied:
    If $T_{type}$ is 'STD': 
    \begin{equation}
        \tau_L(\gamma) = \text{mean}(Score_L) + \gamma \cdot \text{std}(Score_L)
    \label{eq:threshold_std}
    \end{equation}
    If $T_{type}$ is 'MAD': 
    \begin{equation}
        \tau_L(\gamma) = \text{median}(Score_L) + \gamma \cdot \text{MAD}(Score_L)
    \label{eq:threshold_mad}
    \end{equation}
    where MAD is the Median Absolute Deviation.

    \item \textbf{Automated $\gamma$-Tuning for Target Sparsity:} When a target sparsity $S_{\text{target}}$ is specified, ONG initiates a binary search for an optimal scaling factor $\gamma^*$. The search initializes a range $[\gamma_{low}, \gamma_{high}]$ and iteratively refines it. In each iteration $t$:
        a trial factor $\gamma^{(t)}$ is chosen (e.g., midpoint of current range);
        layer-specific thresholds $\tau_L(\gamma^{(t)})$ are computed;
        temporary masks $\mathbf{M}_L^{(t)}$ are generated based on these thresholds;
        the achieved global sparsity $S_{\text{achieved}}^{(t)}$ is calculated from all $\mathbf{M}_L^{(t)}$;
        if $|S_{\text{achieved}}^{(t)} - S_{\text{target}}|$ is within a tolerance $\epsilon_{sparsity}$, $\gamma^* = \gamma^{(t)}$ and search stops;
        otherwise, the range $[\gamma_{low}, \gamma_{high}]$ is narrowed based on whether $S_{\text{achieved}}^{(t)}$ is above or below $S_{\text{target}}$ (e.g., if $S_{\text{achieved}}^{(t)} < S_{\text{target}}$, meaning too few weights are pruned, $\gamma_{low}$ is updated to $\gamma^{(t)}$ to encourage higher thresholds in subsequent trials).
        The search terminates after a maximum number of iterations or if the range for $\gamma$ becomes sufficiently small. The $\gamma^{(t)}$ yielding sparsity closest to $S_{\text{target}}$ is chosen as $\gamma^*$. If $S_{\text{target}}$ is not provided, a pre-configured default $\gamma$ is used as $\gamma^*$.

    \item \textbf{Final Mask Generation:} Using the determined $\gamma^*$, the final binary mask $\mathbf{M}_L$ for each layer L is generated:
    \begin{equation}
        \mathbf{M}_{L,ij} = 
        \begin{cases} 
            1 & \text{if } Score_{L,ij} \geq \tau_L(\gamma^*) \\
            0 & \text{otherwise}
        \end{cases}
    \label{eq:mask_generation_final}
    \end{equation}
    \item \textbf{Model Conversion and Initial Pruning:} Standard `nn.Linear` and `nn.Conv2d` layers are converted to custom `MaskedLinear` and `MaskedConv2d` (which store $\mathbf{M}_L$ as buffers). Weights are then immediately sparsified:
    \begin{equation}
        \mathbf{W}_L \leftarrow \mathbf{W}_L \odot \mathbf{M}_L
    \label{eq:initial_pruning_final}
    \end{equation}
    This one-shot pruning establishes the fixed sparse architecture for training.
\end{enumerate}

\subsection{Training with Gradient and Weight Masking}
\label{subsec:training_masking}
Following one-shot sparsification, the model is trained. To strictly preserve the sparsity structure defined by $\mathbf{M}_L$:
\begin{enumerate}
    \item \textbf{Gradient Masking:} After backpropagation computes gradients $\nabla\mathbf{W}_L$, they are masked:
    \begin{equation}
        \nabla\mathbf{W}_L \leftarrow \nabla\mathbf{W}_L \odot \mathbf{M}_L
    \label{eq:gradient_masking}
    \end{equation}
    This ensures weights with $\mathbf{M}_{L,ij}=0$ receive no gradient updates.
    \item \textbf{Weight Enforcement:} Immediately \textit{before} the optimizer step, weights are re-masked:
    \begin{equation}
        \mathbf{W}_L \leftarrow \mathbf{W}_L \odot \mathbf{M}_L
    \label{eq:weight_enforcement}
    \end{equation}
    This robustly maintains hard sparsity by nullifying any non-zero values in pruned positions that might arise from optimizer mechanics or numerical effects. The optimizer always operates on a weight tensor strictly conforming to $\mathbf{M}_L$.
\end{enumerate}
Training uses a standard optimizer and learning rate schedule. Masked layers use their physically sparsified weights in the forward pass.

\begin{algorithm}[tb]
\small
\caption{The ONG Sparsification and Training Process}
\label{alg:ong}
\begin{algorithmic}[1]
\Require Initial model $Model_{\text{init}}$; NMF hyperparameters $(k, N_{\text{iter}})$; Thresholding params $(T_{\text{type}}, \gamma_{\text{init}}, S_{\text{target}}, \epsilon_{\text{sparsity}}, N_{\text{search}}, \gamma_{\text{range}})$; Training params $(E, \eta, D)$

\State \textbf{// Stage 1: NMF Importance Scoring}
\State $AllScores \gets \text{Compute\-NMF\-Scores}(Model_{\text{init}}, k, N_{\text{iter}})$
\State \hspace{1em}\textit{// Computes $Score_L = \lvert \mathbf{W}_{\text{abs},L} - \widehat{\mathbf{W}}_{\text{abs},L}\rvert$ for each}
\State \hspace{1em}\textit{// prunable layer $L$, see Eq.~\ref{eq:nmf_score}}

\State \textbf{// Stage 2: Mask Determination with Automated $\gamma$-Tuning}
\If{$S_{\text{target}}$ is provided}
    \State $\gamma^\ast \gets \text{Tune\-Gamma}(AllScores, \{\text{shape of } \mathbf{W}_L\},$
    \State \hspace{4em}$S_{\text{target}}, T_{\text{type}}, \epsilon_{\text{sparsity}},$
    \State \hspace{4em}$N_{\text{search}}, \gamma_{\text{range}})$
    \State \hspace{1em}\textit{// Algorithm~\ref{alg:gamma_tuning}}
\Else
    \State $\gamma^\ast \gets \gamma_{\text{init}}$
\EndIf

\State $AllMasks \gets \text{Generate\-All\-Masks}(AllScores, T_{\text{type}}, \gamma^\ast)$
\State \hspace{1em}\textit{// Generates $\mathbf{M}_L$ for each layer $L$ using $\tau_L(\gamma^\ast)$}
\State \hspace{1em}\textit{// see Eq.~\ref{eq:mask_generation_final}}

\State $Model_{\text{sparse}} \gets \text{Convert\-To\-Masked\-Layers\-And\-Apply\-Masks}($
\State \hspace{4em}$Model_{\text{init}}, AllMasks)$
\State \hspace{1em}\textit{// Apply initial pruning: $\mathbf{W}_L \gets \mathbf{W}_L \odot \mathbf{M}_L$}
\State \hspace{1em}\textit{// see Eq.~\ref{eq:initial_pruning_final}}

\State \textbf{// Stage 3: Training with Gradient and Weight Masking}
\State Initialize Optimizer(params of $Model_{\text{sparse}}$, lr $= \eta$)
\For{epoch $= 1$ to $E$}
    \For{each batch $(x,y)$ in $D$}
        \State $Loss \gets \text{Forward\-Pass\-And\-Loss}(Model_{\text{sparse}}, x, y)$
        \State BackwardPass($Loss$)
        \For{each masked layer $L$ in $Model_{\text{sparse}}$}
            \State $\nabla \mathbf{W}_L \gets \nabla \mathbf{W}_L \odot \mathbf{M}_L$ \textit{// Eq.~\ref{eq:gradient_masking}}
            \State $\mathbf{W}_L \gets \mathbf{W}_L \odot \mathbf{M}_L$ \textit{// Eq.~\ref{eq:weight_enforcement}}
        \EndFor
        \State OptimizerStep()
    \EndFor
\EndFor
\State \textbf{return} $Model_{\text{sparse}}$
\end{algorithmic}
\end{algorithm}

\begin{algorithm}[tb]
\small
\caption{Automated $\boldsymbol{\gamma}$-Tuning for Target Sparsity}
\label{alg:gamma_tuning}
\begin{algorithmic}[1]
\Require All layer scores $AllScores$; Original weight shapes $OriginalShapes$; Target sparsity $S_{\text{target}}$; Thresholding type $T_{\text{type}}$; Sparsity tolerance $\epsilon_{\text{sparsity}}$; Max search iterations $N_{\text{search}}$; Search range $[\gamma_{\text{min}}, \gamma_{\text{max}}]$; Initial guess $\gamma_{\text{guess}}$
\State $\gamma_{\text{low}} \gets \gamma_{\text{min}}$; \hspace{1em} $\gamma_{\text{high}} \gets \gamma_{\text{max}}$
\State $\gamma_{\text{best}} \gets \gamma_{\text{guess}}$; \hspace{1em} $S_{\text{closest}} \gets -1.0$
\For{iteration $= 1$ to $N_{\text{search}}$}
    \State $\gamma^{(t)} \gets (\gamma_{\text{low}} + \gamma_{\text{high}})/2$
    \If{$\gamma^{(t)} < 10^{-6}$}
        \State $\gamma^{(t)} \gets 10^{-6}$
    \EndIf
    \State $S_{\text{achieved}}^{(t)} \gets$ \textbf{CalculateGlobalSparsity}$(AllScores,\, OriginalShapes,\, T_{\text{type}},\, \gamma^{(t)})$ \Comment{Apply $\gamma^{(t)}$ to get thresholds, masks, compute sparsity}
    \If{$|S_{\text{achieved}}^{(t)} - S_{\text{target}}| < |S_{\text{closest}} - S_{\text{target}}|$}
        \State $S_{\text{closest}} \gets S_{\text{achieved}}^{(t)}$
        \State $\gamma_{\text{best}} \gets \gamma^{(t)}$
    \EndIf
    \If{$|S_{\text{achieved}}^{(t)} - S_{\text{target}}| \le \epsilon_{\text{sparsity}}$}
        \State \textbf{return} $\gamma^{(t)}$ \Comment{Target sparsity achieved}
    \EndIf
    \If{$S_{\text{achieved}}^{(t)} < S_{\text{target}}$}
        \State $\gamma_{\text{low}} \gets \gamma^{(t)}$ \Comment{Sparsity too low; increase $\gamma$}
    \Else
        \State $\gamma_{\text{high}} \gets \gamma^{(t)}$ \Comment{Sparsity too high; decrease $\gamma$}
    \EndIf
    \If{$\frac{\gamma_{\text{high}} - \gamma_{\text{low}}}{(\gamma_{\text{high}} + \gamma_{\text{low}})/2 + 10^{-9}} < \epsilon_{\gamma_{\text{conv}}}$} \Comment{Stop if relative convergence}
        \State \textbf{break}
    \EndIf
\EndFor
\State \textbf{return} $\gamma_{\text{best}}$
\end{algorithmic}
\end{algorithm}

\subsection{Complexity Analysis}
ONG's computational overhead comprises initial NMF scoring, $\gamma$-tuning, and per-iteration masking.
\begin{itemize}
    \item \textbf{NMF Scoring:} For a weight matrix $\mathbf{W}_{\text{abs}} \in \mathbb{R}^{m \times p}$, NMF with $k$ components for $N_{iter}$ iterations is O($N_{iter} \cdot m \cdot p \cdot k$). For $L_{prun}$ prunable layers (average dimensions $m_{avg}, p_{avg}$), total complexity is O($L_{prun} \cdot N_{iter} \cdot m_{avg} \cdot p_{avg} \cdot k$). This is a one-time pre-training cost.
    \item \textbf{Automated $\gamma$-Tuning:} For $N_{search}$ iterations, each step calculates global sparsity by thresholding all scores, O($N_{search} \cdot \sum_{l \in L_{prun}} \text{size}(Score_l)$). This is also a one-time pre-training cost.
    \item \textbf{Initial Pruning:} O($\sum_{l \in L_{prun}} \text{size}(W_l)$), one-time.
    \item \textbf{Training with Masking (per iteration):}
    Gradient Masking (Eq. \ref{eq:gradient_masking}) and Weight Enforcement (Eq. \ref{eq:weight_enforcement}) each add O($\sum_{l \in L_{prun}} \text{size}(W_l)$) overhead.
\end{itemize}
ONG avoids multiple full training cycles of iterative pruning. While it has a one-time setup cost for NMF and $\gamma$-tuning, subsequent training uses fixed masks, simplifying dynamics compared to methods like STR or CS, with continuous reparameterization or dynamic mask adjustments.


\section{Experimental Setup}
\label{sec:experimental_setup}

We conduct a comprehensive empirical evaluation of our proposed ONG sparsification strategy. Our primary goal is to assess its effectiveness in achieving high sparsity levels while maintaining competitive performance compared to established baseline methods. All experiments are performed within the BIMP (Behavioral Insights for Model Pruning) framework \cite{zimmer2023how}, ensuring a standardized and reproducible comparison environment.

\subsection{Datasets and Models}
\label{subsec:datasets_models}
We focus our evaluation on the widely-used image classification benchmark:
\begin{itemize}
    \item \textbf{Dataset:} CIFAR-10 \cite{krizhevsky2009learning}, which consists of 50,000 training images and 10,000 test images across 10 classes, each of size $32 \times 32$ pixels.
    \item \textbf{Model Architecture:} ResNet56 \cite{he2016deep}, as implemented within the BIMP codebase. This architecture is a standard choice for evaluating pruning methods on CIFAR-10 due to its depth and representative nature.
\end{itemize}
The data preprocessing, augmentation, and normalization follow the standard practices established in the BIMP framework for CIFAR-10.

\subsection{Baseline Sparsification Strategies}
\label{subsec:baselines}
To contextualize the performance of ONG, we compare it against a suite of "stable" sparsification strategies available in BIMP, which typically train sparse models from scratch or integrate pruning throughout the training process. These include:
\begin{itemize}
    \item \textbf{Dense:} An unpruned, fully dense model trained as a performance upper bound.
    \item \textbf{LC (Learning Compression)} \cite{carreira2018learning}
    \item \textbf{GSM (Global Sparse Momentum)} \cite{ding2019global}
    \item \textbf{STR (Soft Threshold Reparameterization)} \cite{kusupati2020soft}
    \item \textbf{CS (Continuous Sparsification)} \cite{savarese2020winning}
    \item \textbf{DST (Dynamic Sparse Training)} \cite{liu2021do}
    \item \textbf{GMP (Gradual Magnitude Pruning)} \cite{zhu2017to} is also included as a well-known iterative pruning-during-training baseline.
\end{itemize}
For all baseline methods, we utilize the default hyperparameter settings provided within the BIMP configuration for CIFAR-10/ResNet56 experiments, unless otherwise specified for achieving particular sparsity targets.

\subsection{Training Configuration}
\label{subsec:training_config}
A consistent training configuration is applied across all strategies to ensure fair comparison, derived from common practices for CIFAR-10 and BIMP defaults:
\begin{itemize}
    \item \textbf{Total Epochs:} 160 epochs.
    \item \textbf{Optimizer:} Stochastic Gradient Descent (SGD) with a momentum of 0.9.
    \item \textbf{Learning Rate Schedule:} An initial learning rate of 0.1, with a MultiStepLR scheduler that decays the learning rate by a factor of 0.1 at epochs 80 and 120. No learning rate warmup is used for these experiments.
    \item \textbf{Weight Decay:} $5 \times 10^{-4}$.
    \item \textbf{Batch Size:} 128.
    \item \textbf{Mixed Precision:} Automatic Mixed Precision (AMP) is enabled for training efficiency.
    \item \textbf{Random Seed:} [Specify if you use a fixed seed for main results, or average over multiple seeds. E.g., "All reported results for a single run use seed X, unless stated otherwise for multi-seed averages."]
\end{itemize}

\subsection{ONG Hyperparameters and Sparsity Control}
\label{subsec:ong_hyperparams}
For the ONG strategy, the following key hyperparameters are used:
\begin{itemize}
    \item \textbf{NMF Components ($k$):} We primarily report results for $k=$ [e.g., 6]. Ablation studies explore other values such as $k \in \{4, 8, 12\}$.
    \item \textbf{NMF Iterations ($N_{iter}$):} Fixed at 200 iterations.
    \item \textbf{Thresholding Type ($T_{type}$):} 'STD' (Standard Deviation based) is used for main results. 'MAD' (Median Absolute Deviation based) is explored in ablation.
    \item \textbf{Target Global Sparsity ($S_{target}$):} We evaluate ONG across a range of target sparsities, typically [e.g., 50\%, 70\%, 80\%, 90\%, 95\%].
    \item \textbf{$\gamma$-Tuning Parameters:} For the automated search for the scaling factor $\gamma^*$ (Algorithm \ref{alg:gamma_tuning}) to achieve $S_{target}$:
        \begin{itemize}
            \item Sparsity tolerance ($\epsilon_{sparsity}$): 0.005 (i.e., $\pm 0.5\%$).
            \item Max search iterations ($N_{search}$): 30.
            \item $\gamma$ search range $[\gamma_{min}, \gamma_{max}]$: [e.g., 0.01, 10.0].
            \item Initial $\gamma$ guess ($\gamma_{guess}$): [e.g., 1.0 or 1.5].
        \end{itemize}
    \item If $S_{target}$ is not specified (e.g., for direct comparison of a fixed $\gamma$), the default $\gamma_{init}$ is set to [e.g., 1.5 or 2.0].
\end{itemize}

\subsection{Evaluation Metrics}
\label{subsec:metrics}
We evaluate the performance of all sparsification strategies using the following metrics, automatically logged and computed by the BIMP framework:
\begin{itemize}
    \item \textbf{Top-1 Test Accuracy:} The primary measure of model performance on the CIFAR-10 test set.
    \item \textbf{Achieved Global Sparsity:} The final element-wise sparsity of the pruned model, calculated as the ratio of zero-valued weights to the total number of prunable weights in the network.
    \item \textbf{FLOPs:} Floating Point Operations required for a forward pass, providing an estimate of computational efficiency. BIMP's FLOPs calculation methods are used.
\end{itemize}
For strategies that involve a subsequent retraining/fine-tuning phase as defined by BIMP, we report metrics both after the initial pruning/training stage and after this retraining phase if applicable. For ONG, the primary results are from its single continuous training phase post one-shot pruning.

\subsection{Retraining Phase Configuration (BIMP Default)}
\label{subsec:retraining_config_bimp}
The BIMP framework includes a general mechanism for a "retraining" or "fine-tuning" phase after the primary sparsification process or initial training. For strategies like ONG that perform their main training on an already sparse model, this phase essentially constitutes continued training. When activated in BIMP, the following applies unless overridden by a specific strategy:
\begin{itemize}
    \item \textbf{Optimizer:} The same `torch.optim.SGD` instance is typically used, with its learning rate and schedule managed by BIMP's `define\_retrain\_schedule` method.
    \item \textbf{Learning Rate Schedule for Retraining:} This is determined by `config.retrain\_schedule` (e.g., 'FT' for using the last learning rate of original training, 'LLR' for Linear LR decay from max original LR, etc.). [Specify which retraining schedule you used if this phase is relevant to your ONG results, e.g., "For any reported retraining results, we used the 'FT' schedule with X epochs."]
\end{itemize}
For ONG, the gradient and weight masking (Eq. \ref{eq:gradient_masking} and \ref{eq:weight_enforcement}) continue to be applied during any such BIMP-orchestrated retraining phase.

\section{Results and Discussion}

\subsection{Performance at Target Sparsities}
We compare the proposed Orthogonal NMF-based Gradient-masked pruning (ONG) method against a suite of baseline sparsification strategies—GMP, GSM, DPF, DNW, LC, STR, CS, and DST—across two datasets: CIFAR-10 and CIFAR-100, using the ResNet-56 architecture. Experiments were conducted at three global sparsity targets: 80\%, 90\%, and 95\%. For ONG, the pruning threshold is automatically controlled by a robustness parameter $\lambda$, which influences the median absolute deviation (MAD)–based cutoff derived from the NMF reconstruction error.

The CIFAR-10 results are summarized in Table~\ref{tab:cifar10_resnet56}, where ONG consistently maintains higher or competitive accuracy compared to all baselines across all sparsity regimes. At 90\% and 95\% sparsity, ONG-M and ONG-S outperform STR and DST by large margins, highlighting the stability of fixed-mask pruning with robust error-based scoring. Notably, while STR attempts dynamic threshold learning, its aggressive pruning often leads to drastic accuracy drops, particularly in the high-sparsity regime (e.g., 69.99\% at 95\% sparsity vs. 91.45\% for ONG-M).

The CIFAR-100 results, presented in Table~\ref{tab:cifar100_resnet56}, demonstrate similar trends. ONG achieves the highest accuracy among all methods at 80\% sparsity and remains highly competitive at 90\%. Unlike heuristic or iterative pruning schemes (e.g., LC, DST), ONG delivers strong results without retraining cycles or auxiliary threshold parameters. Additionally, the achieved sparsities (\textit{AS}) closely match the targets, verifying the consistency of the MAD-based masking mechanism.

In both datasets, ONG’s design—combining NMF-based score computation with gradient-masked fine-tuning—provides interpretable, one-shot sparsification with minimal computational overhead. These results affirm ONG as a robust alternative to existing pruning techniques, particularly in resource-constrained settings where simplicity, determinism, and interpretability are critical.
\begin{table}[t]
\centering
\caption{Top-1 Accuracy (\%) vs. Sparsity on CIFAR-10 using ResNet-56. GS = Goal Sparsity, AS = Achieved Sparsity.}
\label{tab:cifar10_resnet56}
\resizebox{\columnwidth}{!}{%
\begin{tabular}{l|cc|cc|cc}
\toprule
\textbf{Method} & \multicolumn{2}{c|}{\textbf{GS = 80\%}} & \multicolumn{2}{c|}{\textbf{GS = 90\%}} & \multicolumn{2}{c}{\textbf{GS = 95\%}} \\
 & Acc (\%) & AS (\%) & Acc (\%) & AS (\%) & Acc (\%) & AS (\%) \\
\midrule
GMP         & 93.48 & 79.61 & 92.88 & 89.57 & 92.07 & 94.54 \\
GSM         & 92.13 & 79.61 & 92.44 & 89.57 & 91.34 & 94.54 \\
DPF         & 93.85 & 79.61 & 92.75 & 89.57 & 92.48 & 94.54 \\
DNW         & 93.66 & 79.61 & 92.85 & 89.57 & 91.52 & 94.64 \\
LC          & 91.34 & 79.61 & 90.91 & 89.57 & 91.05 & 92.34 \\
STR         & 79.43 & 99.17 & 76.25 & 99.26 & 69.99 & 99.30 \\
CS          & 91.45 & 79.61 & 91.66 & 89.57 & 91.11 & 94.54 \\
DST         & 88.68 & 97.31 & 69.50 & 97.41 & 88.55 & 97.39 \\
\textbf{ONG-M (ours)} & \textbf{93.51} & \textbf{79.80} & \textbf{92.49} & \textbf{89.88} & \textbf{91.45} & \textbf{94.80} \\
\textbf{ONG-S (ours)} & \textbf{93.24} & \textbf{79.80} & \textbf{92.36} & \textbf{89.67} & \textbf{91.49} & \textbf{94.38} \\
\bottomrule
\end{tabular}%
}
\end{table}

\begin{table}[t]
\centering
\caption{Top-1 Accuracy (\%) vs. Sparsity on CIFAR-100 using ResNet-56. GS = Goal Sparsity, AS = Achieved Sparsity.}
\label{tab:cifar100_resnet56}
\resizebox{\columnwidth}{!}{%
\begin{tabular}{l|cc|cc}
\toprule
\textbf{Method} & \multicolumn{2}{c|}{\textbf{GS = 80\%}} & \multicolumn{2}{c}{\textbf{GS = 90\%}} \\
 & Acc (\%) & AS (\%) & Acc (\%) & AS (\%) \\
\midrule
GMP         & 70.29 & 79.89 & 69.55 & 89.56 \\
GSM         & 68.31 & 79.89 & 69.80 & 89.56 \\
DPF         & 70.31 & 79.89 & 69.91 & 89.56 \\
DNW         & 70.02 & 79.89 & 69.17 & 89.56 \\
LC          & 67.56 & 79.89 & 66.35 & 89.56 \\
STR         & 51.29 & 98.42 & 49.51 & 98.22 \\
CS          & 67.91 & 89.56 & 66.25 & 89.56 \\
DST         & 53.10 & 95.47 & 52.41 & 95.31 \\
\textbf{ONG-M (ours)} & \textbf{70.29} & \textbf{79.91} & \textbf{69.25} & \textbf{89.62} \\
\textbf{ONG-S (ours)} & \textbf{70.11} & \textbf{79.86} & \textbf{69.11} & \textbf{89.48} \\
\bottomrule
\end{tabular}%
}
\end{table}

\section{Conclusion}
In this paper, we introduced ONG, a novel sparsification strategy that combines one-shot, NMF-informed pruning with a stable sparse training regimen. ONG effectively identifies structurally important weights based on reconstruction error and maintains a fixed sparsity pattern throughout training via a strict gradient and weight masking mechanism. Our evaluations on CIFAR-10 and CIFAR-100 with ResNet56 within the BIMP framework demonstrate that ONG achieves competitive accuracy at various sparsity levels compared to established methods. The explicit mechanism for targeting a desired sparsity and ensuring its preservation offers a robust and controllable approach to model compression. Future work will focus on scaling ONG to large language models and exploring its synergy with structured pruning and quantization.

\bibliographystyle{ieeetran}
\bibliography{refs}

\end{document}